\documentclass[letterpaper, 10 pt, conference]{ieeeconf}  %

\IEEEoverridecommandlockouts                              %

\usepackage{graphics} %
\usepackage{graphicx}

\usepackage[font=small]{caption} 
\usepackage[labelformat=simple]{subcaption}
\usepackage{fixltx2e}
\usepackage{amsmath} %
\usepackage{amssymb}  %
\usepackage{cite}
\usepackage{array}
\usepackage{makecell}
\usepackage{hyperref}
\usepackage{soul}
\usepackage{xcolor}
\usepackage{balance}

\title{SocRATES:  Towards Automated Scenario-based Testing of\\Social Navigation Algorithms}
\author{Shashank Rao Marpally $^{1}$, Pranav Goyal$^{1,2}$, and Harold Soh$^{1,2}$\\$^{1}$Dept. of Computer Science, National University of Singapore.
\\ $^{2}$Smart Systems Institute, NUS.
\\{\small Email: \texttt{\{smarpall@comp.nus.edu.sg, prgoyal@umich.edu, harold@comp.nus.edu.sg\}}}
}

\begin{document}

\maketitle
\thispagestyle{empty}
\pagestyle{empty}

\begin{abstract}

Current social navigation methods and benchmarks primarily focus on proxemics and task efficiency. While these factors are important, qualitative aspects --- such as perceptions of a robot's social competence --- are equally crucial for successful adoption and integration into human environments. We propose a more comprehensive evaluation of social navigation through scenario-based testing, where specific human-robot interaction scenarios can reveal key robot behaviors. However, creating such scenarios is often labor-intensive and complex. In this work, we address this challenge by introducing a pipeline that automates the generation of context-, and location-appropriate social navigation scenarios, ready for simulation. Our pipeline transforms simple scenario metadata into detailed textual scenarios, infers pedestrian and robot trajectories, and simulates pedestrian behaviors, which enables more controlled evaluation. We leverage the social reasoning and code-generation capabilities of Large Language Models (LLMs) to streamline scenario generation and translation. Our experiments show that our pipeline produces realistic scenarios and significantly improves scenario translation over naive LLM prompting. Additionally, we present initial feedback from a usability study with social navigation experts and a case-study demonstrating a scenario-based evaluation of three navigation algorithms.

\end{abstract}

\section{Introduction}

  \setlength{\abovecaptionskip}{1ex}
 \setlength{\belowcaptionskip}{1ex}
 \setlength{\floatsep}{1ex}
 \setlength{\textfloatsep}{1ex}

In social navigation, robots must navigate through dynamic human environments while adhering to social norms. This presents a dual challenge: ensuring both competent navigation and socially appropriate behavior. Evaluating these aspects is difficult, as social appropriateness is subjective and context-dependent, often requiring human judgment. Current evaluation methods tend to focus on quantifiable metrics like safety and comfort, but these fail to capture the full spectrum of social interactions. If we could automate the generation of socially relevant scenarios, grounded in both location and task, we would significantly enhance the ability to test and refine social navigation algorithms. This would lead to more reliable and robust robots that can function effectively in real-world human environments.

\begin{figure}
    \centering
    \includegraphics[width=1\linewidth]{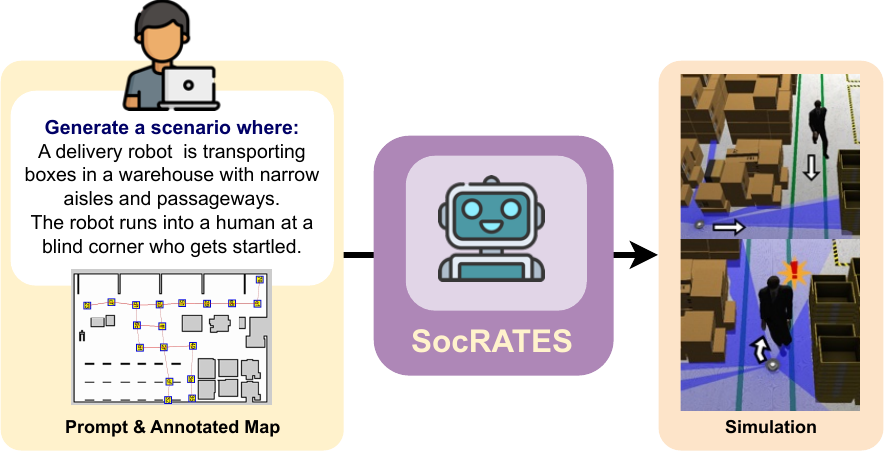}
    \caption{We propose SocRATES, an automated system that leverages VLMs to generate simulated social navigation scenarios from simple textual and image inputs.}
    \label{fig:inputoutput}
\end{figure}

In this work, we propose SocRATES (\textbf{Soc}ial \textbf{R}obot \textbf{A}ssessment \textbf{T}hrough \textbf{S}cenario \textbf{E}valuation) (Fig. \ref{fig:inputoutput}), a system designed to automate the generation of context-specific scenarios for evaluating social navigation algorithms. The key insight behind SocRATES is the use of large  vision-language models (VLMs) to translate high-level, potentially ambiguous inputs into detailed simulation scenarios. By leveraging the commonsense reasoning and context-understanding capabilities of these models, SocRATES generates realistic and varied scenarios that capture both the physical layout of the environment and the social interactions that occur within it. We believe this combination of AI-driven scenario generation with simulation makes SocRATES a significant first-step towards testing both the navigation and social appropriateness of robots.

Our work builds on prior research in scenario-based testing for autonomous systems and social navigation. Tools like SEAN 2.0 \cite{tsoi2022sean} and HuNavSim \cite{perez2023hunavsim} have explored human behavior simulation in social settings, but they remain limited by a narrow set of predefined scenarios and/or lack flexibility or control over scenario generation. In contrast, SocRATES allows users to define scenarios based on textual and image-based inputs, providing much greater control over the evaluation process. Additionally, while prior tools focus primarily on the fidelity of human simulation, SocRATES is designed to evaluate the robot's behavior in relation to both the task and social context. To our knowledge, SocRATES is the first system to integrate LLM-driven scenario generation with simulation for the comprehensive testing of social navigation algorithms.

Our experimental results demonstrate the effectiveness of SocRATES in generating diverse, contextually relevant scenarios. We conducted a design analysis that shows the system is fast and cost-effective, producing scenarios in under a minute with minimal cost. A user study with researchers in social navigation validated the system's usability, with all participants preferring SocRATES over manual scenario generation. Ablation studies further highlighted the importance of structured prompts and error-handling mechanisms in ensuring the quality of generated scenarios. Finally, we established a human study benchmark to compare multiple navigation algorithms, showcasing the practical utility of SocRATES as an evaluation tool.

To summarize, the key contributions of this work are:
\begin{itemize}
    \item A novel system for automating the generation of social navigation scenarios using LLMs and VLMs. SocRATES uses   commonsense reasoning and contextual understanding to create realistic, task-relevant scenarios.
    \item A flexible framework that allows users to define scenarios based on high-level inputs, providing control over evaluation scenarios.
    \item Experimental validation through user studies and benchmarks, demonstrating the system's utility and efficiency.
\end{itemize}
By enabling automated, scalable testing of both navigation and social appropriateness, SocRATES moves the field closer to developing robots that can reliably and effectively navigate human environments.

 \begin{figure*}
    \centering
    \includegraphics[width=1\linewidth]{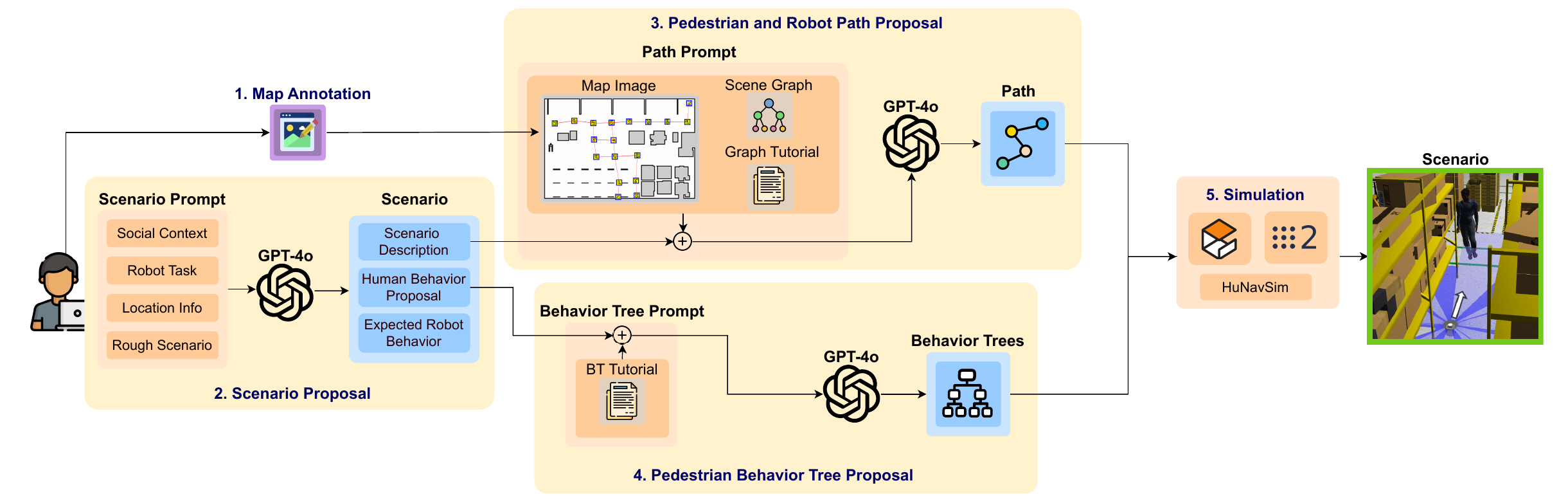}
    
\caption{Overview of our pipeline. We prompt users to annotate the map of their desired location (1) and provide simple textual inputs for their desired scenario. Our pipeline proposes a scenario (2) and then generates the 2 main components of the scenario with structured prompts to an LLM: The paths of the robot and pedestrians (3) and the behavior of the humans (4). Finally these are used by the HuNavSim\cite{perez2023hunavsim} framework (5) to generate a simulation of the scenario.}
    \label{pipelinechart}
\end{figure*}

 \section{Background}
 
 \subsection{Evaluating Social Navigation}
Evaluating social navigation is complex because it involves two interlinked components: social appropriateness and competent navigation. Comprehensive evaluation requires assessing both, yet the subjective nature of social appropriateness makes it particularly challenging~\cite{mavrogiannis2023core}. For example, \cite{francis2023principles} outlines seven key principles that robots should follow to be considered socially appropriate. Most current evaluation methods primarily test two of these principles: safety (P1) and comfort (P2), often using proxemics-based metrics that measure the robot’s and human’s relative positions and speeds~\cite{hall1966hidden}. However, principles like contextual appropriateness (P8) depend on the robot's social context and task, which are often overlooked due to their qualitative nature.

To address these challenges, recent approaches have emerged in two main directions. The first uses proxy metrics that model specific aspects of a robot’s social awareness. For instance, \cite{okal2016formalizing} introduces personal and group space intrusion metrics, which are subsequently adopted in other algorithms \cite{perez2018teaching} and evaluation frameworks \cite{perez2023hunavsim}. The second approach involves using human studies to assess social behaviors. For example, legibility has been modeled~\cite{dragan2013legibility} and used to improve social navigation. Metrics like Envelope of Readiness, Clarity, and Moments of Confusion~\cite{taylor2022observer} rely on continuous observer data to quantify legibility. Additionally, prior studies~\cite{honour2021perceived} collect qualitative perceptions of social navigation using validated scales, e.g., PSI~\cite{barchard2020measuring}.

However, collecting scalable and informative human study data remains challenging, and existing benchmarks provide limited coverage of edge cases and special scenarios~\cite{francis2023principles}. Crafting specific simulations for testing edge cases is cumbersome, contributing to the difficulty in gathering useful human evaluation data. These limitations highlight the need for more comprehensive scenario-based testing, where specific behavioral aspects of the robot can be evaluated in controlled settings. Our work addresses this gap by automating the proposal and generation of context- and location-specific scenarios from simple textual and image-based inputs.

\subsection{Scenario-based Testing for Social Navigation}

Automated scenario generation for testing has a long history in Advanced Driver Assistance Systems (ADAS) \cite{ma2021traffic,wen2020scenario,zhong2023language}, with recent efforts extending into Human-Robot Interaction (HRI) policy evaluation~\cite{2022fontaine,bhatt2023surrogate}. However, most current social navigation evaluation tools, such as HuNavSim~\cite{perez2023hunavsim}, Arena~\cite{kästner2024arena}, and SocNavBench~\cite{biswas2022socnavbench}, focus primarily on the fidelity of human simulation rather than generating specific scenarios that test a robot's navigation and reactive capabilities.

SEAN 2.0 \cite{tsoi2022sean} is one exception that seeks to expose the robot to various social situations. However, SEAN 2.0 models only a limited set of scenarios and environments, offering little control over the scenario itself, which increases the uncertainty of whether a specific user-defined scenario will occur during evaluation. To the best of our knowledge, none of the current tools provide comprehensive simulation of human-robot interactions.

In contrast, SocRATES leverages the rich context and commonsense reasoning capabilities of large language models (LLMs), along with a flexible scenario and map parameterization, to generate a diverse set of scenarios, including human-robot interaction scenarios, in any user-defined location. This approach enables more comprehensive testing by allowing the user to specify detailed parameters, ensuring that specific scenarios of interest can be generated and evaluated.

 \section{Methodology}
Our objective is to develop an easy-to-use and reliable system for automating scenario proposal and generation in social navigation testing. Users provide simple textual and image-based inputs to guide the scenario proposal, which our system then converts into a simulation. Once the simulation is generated, any robot-planner combination based on the ROS2 Navigation stack\footnote{\url{https://docs.nav2.org}} can be used for evaluation.

Figure~\ref{pipelinechart} provides an overview of our framework. We structure our system by deconstructing a scenario into five main components: (1) the \textit{location} where the scenario takes place (Sec.~\ref{scenegraphgen}), (2) a detailed \textit{description} of the scenario (Sec.~\ref{scenarioproposal}), (3) the \textit{paths} for pedestrians and the robot (Sec.~\ref{trajgensection}), (4) the \textit{behavior} of the pedestrians (Sec.~\ref{behavgensection}), and (5) the \textit{simulation} that realizes these components (Sec.~\ref{siminstantiation}).
Each module of SocRATES interactively queries a Vision-Language Model (VLM) to generate elements of the scenario content. The simulation module then executes the scenario using these generated components and the specified navigation algorithm on the robot.

\subsection{Map Annotation}\label{scenegraphgen}

SocRATES includes a simple map annotation tool that allows users to provide the necessary contextual information about their desired location. This enables users to generate scenarios in custom simulated environments. 

We represent a location as a 2D semantic scene graph, which contains both semantic and spatial information. An example scene graph overlaid on the map image for the Small Amazon Warehouse Gazebo world is shown in Fig.~\ref{pipelinechart}. Our annotation tool guides users in annotating a location map by adding nodes and edges. The schema for this graph can be defined by the user, with the requirement being that the node and edge types are self-explanatory. For example, edges are associated with semantic types (e.g., `intersection' and `hallway').

The scene graph serves two main purposes: (a) it provides both semantic and spatial context regarding different areas in the map and how they are connected, and (b) the structure of the scene graph helps to identify and validate paths for pedestrians and the robot, as explained further in Sec.~\ref{trajgensection}.

\subsection{Scenario Proposal}\label{scenarioproposal}
The scenario proposal module is responsible for translating the user's potentially ambiguous textual inputs into a detailed and precise characterization of the scenario. Inspired by the concept of scenario-cards~\cite{francis2023principles}, we ask users to input  \textit{scenario metadata}: (a) the \textit{social context} in which the robot operates (e.g., ``A quiet old-age home"), (b) the robot's intended \textit{task} (e.g., ``Deliver coffee from the kitchen to the Hall"), (c) an optional \textit{rough scenario} where the user can describe more scenario details to constrain the system's outputs, and (d) a description of the \textit{location}.

We generate a detailed scenario description grounded in the user's provided location by prompting a Large Language Model (LLM) with the scenario metadata, context about the social navigation task, and the capabilities of the pedestrians in the simulator (e.g., \textit{they cannot manipulate objects, only navigate}). 

From the Vision-Language Model (VLM), we extract the following outputs: a precise \textit{scenario description} to condition path and behavior generation, a description of the \textit{behaviors} of humans in the scenario, and the \textit{expected behavior of the robot} when navigating the scenario, which serves as ground truth for evaluation.
We observe that providing example inputs and corresponding scenarios in the prompt significantly improves the quality of the LLM's responses. Therefore, we include a set of handcrafted examples in the prompt to guide the model's output.

\subsection{Pedestrian and Robot Path Generation}\label{trajgensection}
This module generates paths for both pedestrians (with group assignments, if applicable) and the robot, orchestrating their movements to align with the scenario. To ensure consistency with the user's provided location, the path generation is conditioned on the scene graph from Sec.~\ref{scenegraphgen}. Human and robot paths are represented as trajectories on this graph. The VLM is queried with a structured prompt, which includes: the scene graph and the scene-graph annotated map image, pedagogical examples on interpreting scene graphs and path generation, and the scenario description (from Sec.~\ref{scenarioproposal}).

We extract paths for each pedestrian and the robot simultaneously, along with group assignments. We identify specific nodes where the robot encounters each pedestrian, which is useful for timing pedestrian motion in the simulation to ensure the scenario unfolds as intended. A common error by the VLM is the assignment of discontinuous paths on the scene graph. We detect such errors and re-query the LLM when necessary. Despite structured prompts, the VLM may still generate a valid but unsatisfactory path. To address this, this module is interactive, allowing the user to accept, reject, or edit paths using natural language commands (e.g., ``Make the robot's path longer''), which re-queries the LLM for updates.

\subsection{Pedestrian Behavior Generation}\label{behavgensection}

The behavior generation module translates the behavior descriptions from the scenario proposal into encodings compatible with the simulator. We model pedestrians using HuNavSim \cite{perez2023hunavsim}, which supports rich, programmable reactive behaviors through behavior trees (BTs). For each pedestrian, we construct an LLM prompt that includes the \textit{BT node library} (a list of available behavior tree nodes), the required human behaviors (from Sec.~\ref{scenarioproposal}), and pedagogical examples on BT syntax and design rules.

The LLM outputs a behavior tree in XML for each pedestrian, which can be directly imported into HuNavSim to control their actions. However, the default behavior trees in HuNavSim do not support interaction between humans and the robot or complex behaviors. To enable interactive scenarios, we implemented additional behaviors allowing humans to gesture towards other agents, as well as recognize gestures. For example, a simulated human can wait until the robot makes a specific gesture or until another pedestrian performs an action.

\subsection{Simulation}\label{siminstantiation}
This module integrates the generated behaviors and trajectories into a simulation environment. The scene graph paths for each pedestrian are transformed into simulator world frame paths using the map parameters and are used as navigation goals for pedestrians and waypoints for the robot planner. Running the simulation through HuNavSim using ROS2 creates a Gazebo instance where the pedestrians follow the specified paths and act according to the behavior trees. 

For interactions (e.g., gesturing), pedestrians and the robot publish integer-coded gestures to specific ROS2 topics. In our experiments, we found that timing the arrival of pedestrians and the robot at their interaction points is critical for orchestrating the scenario. Since the robot's motion is fully controlled by its onboard navigation algorithm, we introduce a \textit{scenario manager} ROS2 node that synchronizes pedestrian movements with the robot's location, ensuring their timely arrival at interaction points, thus increasing the likelihood that the desired scenario plays out correctly.

 \section{Experiments and Results}
 
We evaluated the effectiveness of SocRATES, by analyzing its reliability, cost, and design choices. We evaluated its utility as a tool through a usability study with academic researchers and separately with a  persona-based assessment.

\subsection{Design Analysis}

We assessed our system for \textit{reliability}, \textit{cost} and \textit{speed}. In addition, we conducted an analysis to validate SocRATES structured design:
\begin{itemize}
\item \textbf{Reliability.} We assessed the reliability of SocRATES for both unguided (no rough scenario) and guided (with rough scenario) generation across two different maps. For unguided generation, we created 20 scenarios in each location, while for guided generation, we generated 5 scenarios, 10 times each, in two locations (100 total). Each scenario was manually evaluated for simulability, contextual appropriateness, and alignment with the input rough scenario. 
We observed a 55\% success rate for unguided generation and a 73\% success rate for guided generation. Most failures were due to incorrect behavior trees --- often caused by the LLM using incorrect node selection or ordering --- and poor trajectories that were too short or inaccurate for the proposed scenario. Note that in this experiment, we did not use interactivity and accepted the first syntactically valid response. It was possible to correct the trajectory-related errors by re-querying the LLM with specific instructions. 

\item \textbf{Cost.} Given that closed-source large models charge based on token usage, we evaluated the total cost of running our system by measuring the combined token length of all prompts, including typical map image inputs (we used the warehouse map). Assuming we accepted the first response for each module, the total input tokens  amounted to approximately 15,000 tokens, which costs less than 30 cents per trial using GPT-4o. 

\item \textbf{Speed.} To assess the time required to generate a typical scenario, we created 20 scenarios in one location (without interactivity). We measured the wall-clock time needed to generate a scenario. On average, it took 43 seconds to generate a scenario, which is considerably faster than manually-crafting and coding scenarios.

\item \textbf{Structured Prompting.}
To evaluate the importance of our structured prompts and error-handling mechanisms, we compared our system to a simpler, naive system for path and behavior generation. The naive system lacked our context-rich prompts and directly queried a VLM for path and behavior proposals. As in the reliability test, we again generated and evaluated 100 guided scenarios and 20 unguided scenarios with the naive approach (with the same social context, task and scenario inputs as SocRATES). We observed a 30\% overall success rate for guided generation and 10\% success rate for unguided generation. This result highlights the critical role of structured prompts and error-handling mechanisms in improving the reliability and quality of scenario generation.

\end{itemize}

In summary, SocRATES is cost-effective and efficient, generating scenarios in under a minute with minimal cost. The system achieves a 73\% success rate for first-cut generation, with improvement through interactive adjustments. The comparative study with the naive system confirms that structured prompts and error-handling are key to improving accuracy. 

\begin{figure*}[!t]
    \centering
    \includegraphics[width=1.0\linewidth]{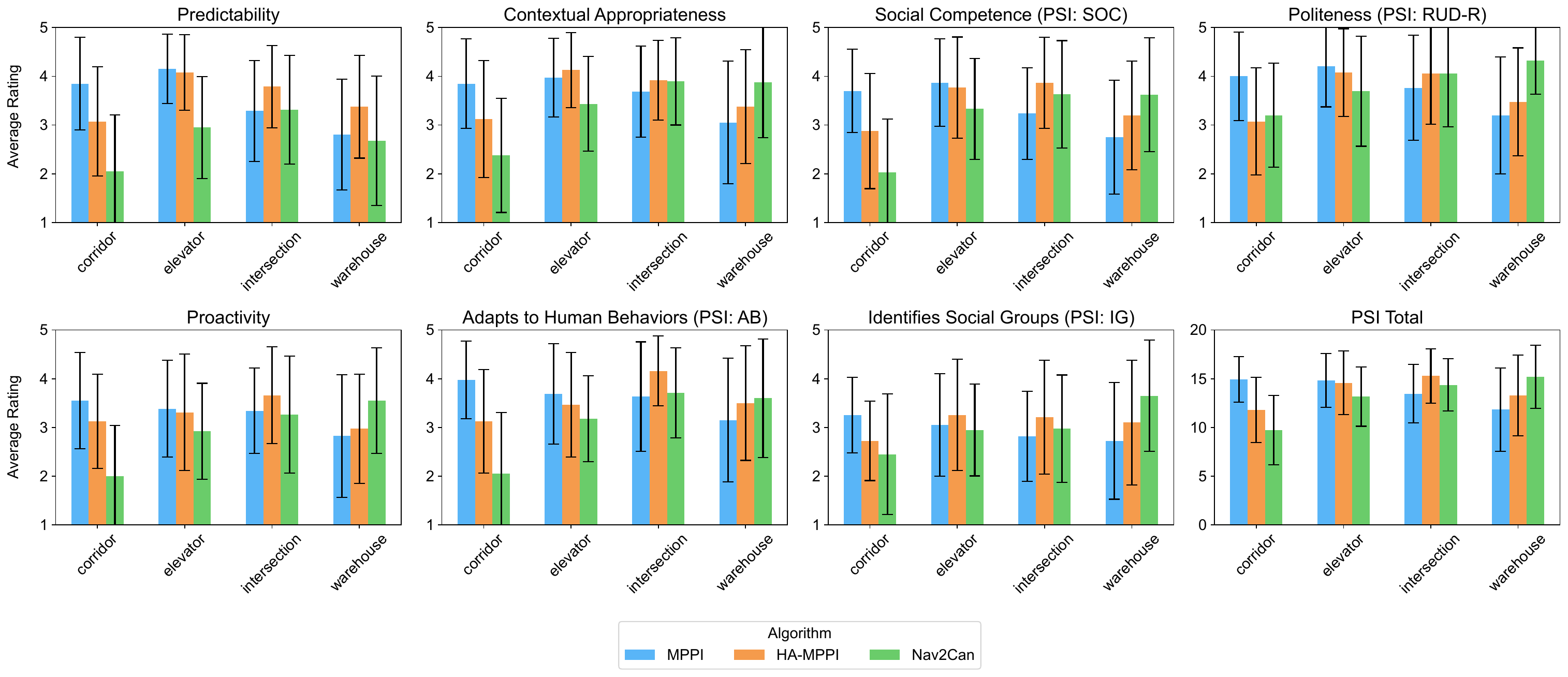}
    \caption{Participant ratings for the navigation algorithms for the four scenarios across various social dimensions.}
    \label{fig:qualtrics_scores}
\end{figure*}

\subsection{Usability Study with Academic Researchers}

We conducted a user study to better understand the usefulness of SocRATES as an evaluation tool for social navigation researchers. We engaged five researchers from the National University of Singapore, Yale University, Miraikan, and LAAS-CNRS. Each participant provided consent\footnote{Our user studies and benchmark were approved under the IRB study no. NUS-IRB-2024-590}, and we demonstrated the map generation process before asking them to input their own scenario metadata. We then showcased the resulting simulation by either teleoperating the robot or running a Nav2-based planner. Below, we summarize the key results:

\begin{itemize}
    \item \textbf{Ease of Use and Reliability:} SocRATES successfully generated the required scenarios for 4 out of 5 users. For one user, the behaviors were semantically correct but syntactically incorrect, an issue that has since been resolved. All participants agreed that the system was easy to use and intuitive. Users expressed a preference for using SocRATES over manually creating scenarios, especially when generating diverse scenarios from ambiguous inputs. However, one user noted that for highly specific scenarios, manual generation may still be preferred.
    
    \item \textbf{Utility:} Users found various aspects of SocRATES valuable as evidenced by feedback: \textit{“It's great that I don’t have to program each human in the scenario by hand.”} and \textit{“Being able to generate scenarios with interaction is useful.”}. Users found that \textit{“Proposing scenarios from potentially ambiguous context and task information is very useful.”}, and \textit{“The variability in LLM-generated human behaviors models real-life randomness in human actions.”} Overall, users found SocRATES beneficial for scenario generation and preferred it over manual scenario development.
    
    \item \textbf{Suggestions:} Users had different use cases and provided a number of suggestions for future improvements. These included integrating more simulators (e.g., Flatland\footnote{\url{https://flatland-simulator.readthedocs.io/en/latest/}} and Isaac\footnote{\url{https://developer.nvidia.com/isaac/sim}}), introducing plug-and-play human models, enhancing the UX design and legibility of the CLI tool, enabling larger crowd simulations, and developing a method to objectively evaluate the fidelity of generated scenarios. These suggestions point to promising directions for future research and development.
\end{itemize}

Overall, this user study demonstrates that SocRATES is a valuable tool for simplifying scenario generation for social navigation algorithms.

\subsection{Case Study via Persona-based Assesssment}

In this section, we demonstrate the usability of SocRATES through a case study and aim to identify potential areas for improvement.

We employed a methodology similar to persona-based assessment~\cite{faily2011persona}, adopting the role of Amy, an HRI researcher. Amy works at a delivery robot  company and is interested in understanding how people perceive the behavior of navigation algorithms in social settings. She has a background in Human-Computer Interaction and is familiar with software development and ROS but does not program robots daily. Amy’s task is to advise her development team on the social capabilities of three different navigation algorithms:
\begin{itemize}
    \item Model Predictive Path Integral Controller (MPPI) with the NavFn Planner\footnote{\url{https://docs.nav2.org/configuration/packages/configuring-navfn.html}};
    \item Human-aware NavFn-MPPI (HA-MPPI), which augments the above with costmaps~\cite{lu2014layered} that penalize entering the social zone of nearby humans;
    \item Nav2Can~\cite{schworer2023nav2can}, which uses costmaps to penalize passing through groups and entering the social zone of nearby humans.
\end{itemize}

Amy's goal is to determine if there is a socially preferred algorithm and how each performs in specific scenarios. Her task involves:
\begin{itemize}
    \item Using SocRATES to create four distinct scenarios and generate videos of the robot’s behavior;
    \item Collecting feedback from an online pool of participants;
    \item Analyzing and reporting the results.
\end{itemize}

We generated four scenarios in which the robot had to:
\begin{enumerate}
    \item Enter an elevator in a multi-floor hospital;
    \item Navigate a corridor in an office environment\footnote{\url{https://github.com/Arena-Rosnav/arena-simulation-setup/tree/master/worlds/arena\_nus\_com1\_building}} with humans following and obstructing its path;
    \item Navigate an intersection where humans suddenly appear from a room in an office;
    \item Navigate past a group of employees in a warehouse.
\end{enumerate}

Adopting the Amy persona, we found scenario generation using SocRATES to be straightforward. For example, to generate the elevator scenario, Amy used the prompt: \textit{The Robot approaches an elevator to go to a different floor to deliver supplies. The elevator opens and has a few people in it. 2 people leave the elevator while one of them is startled by the robot and keeps looking at it and doesn't leave the elevator}. The complete prompts used are available in our online repository\footnote{Repository available after review.}. On average, each scenario took 10 minutes to complete. The most difficult scenario to craft was the group scenario in the warehouse, where the generated trajectories were suboptimal in that the humans were placed in an unrealistic group formation and specific behaviors of the humans were triggered too early. As such, it was necessary to modify the trajectories and behaviors manually (though, this was easier than creating the trajectories from scratch). Annotating the hospital map also required finding a balance between very specific trajectories and giving the navigation planner freedom to choose alternate paths. These observations highlighted areas for improvement in future versions of SocRATES.

\begin{figure}
    \centering
    \begin{subfigure}{0.47\textwidth}
        \includegraphics[width=0.32\textwidth]{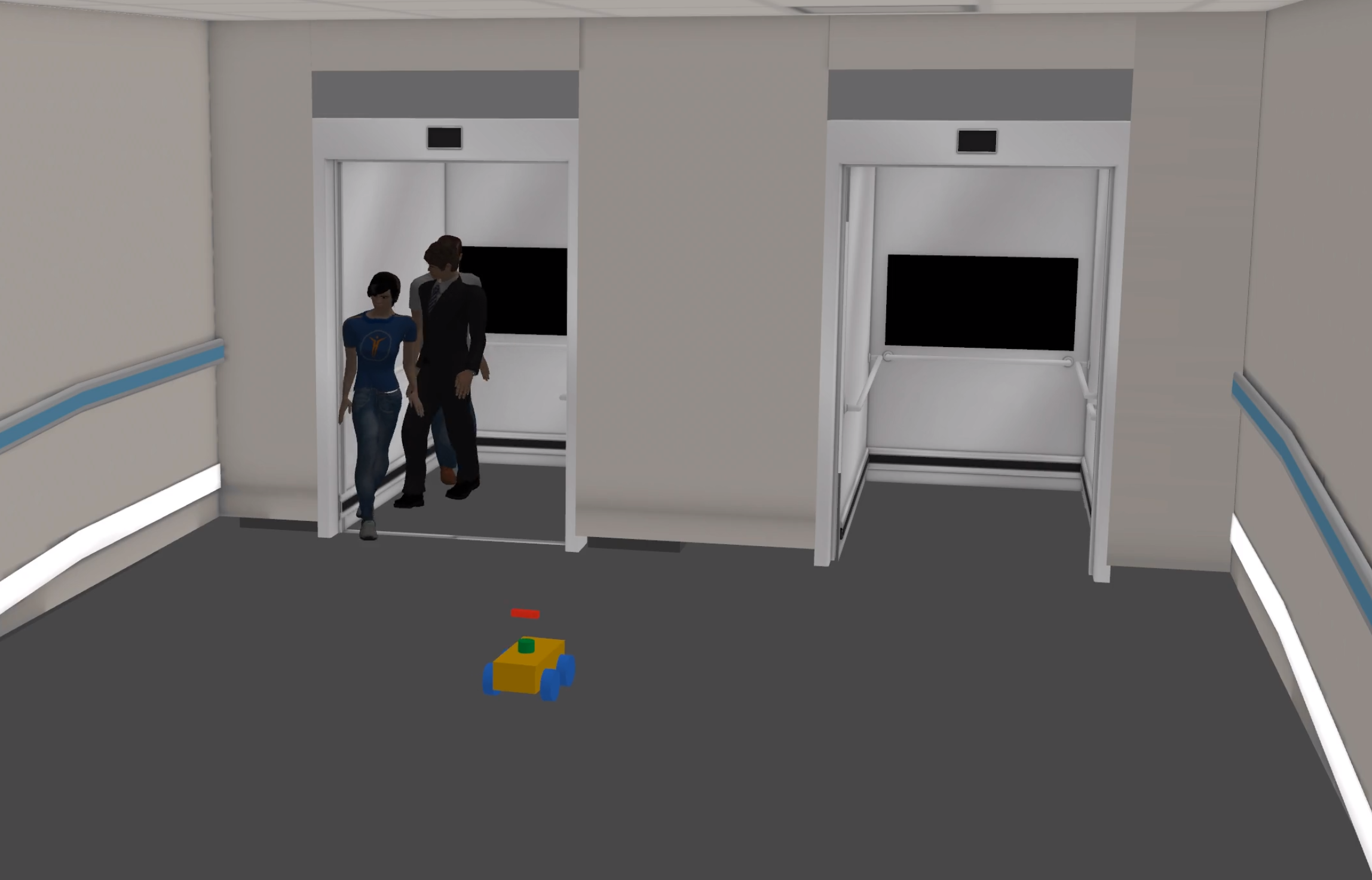}
        \includegraphics[width=0.32\textwidth]{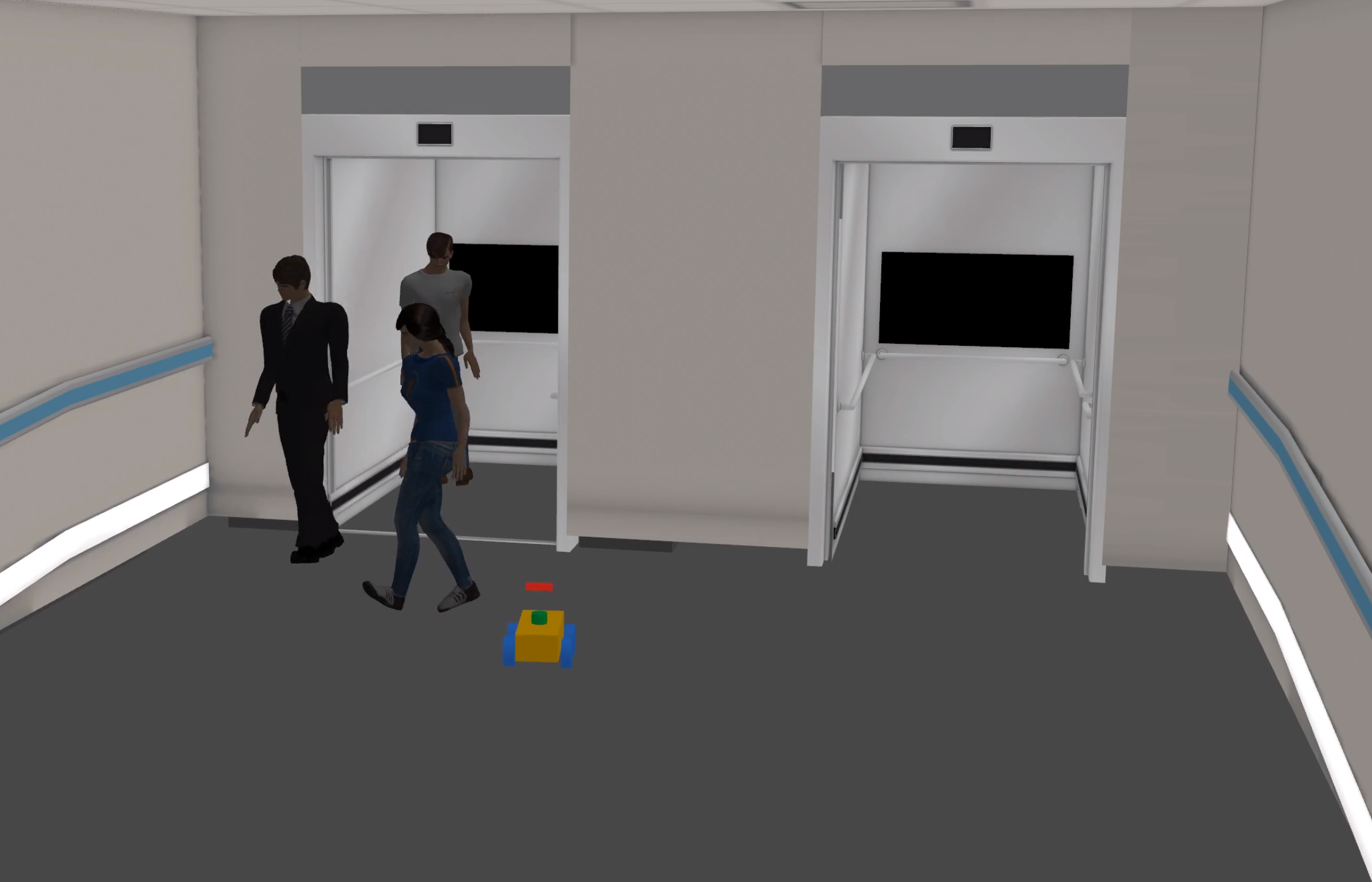}
        \includegraphics[width=0.32\textwidth]{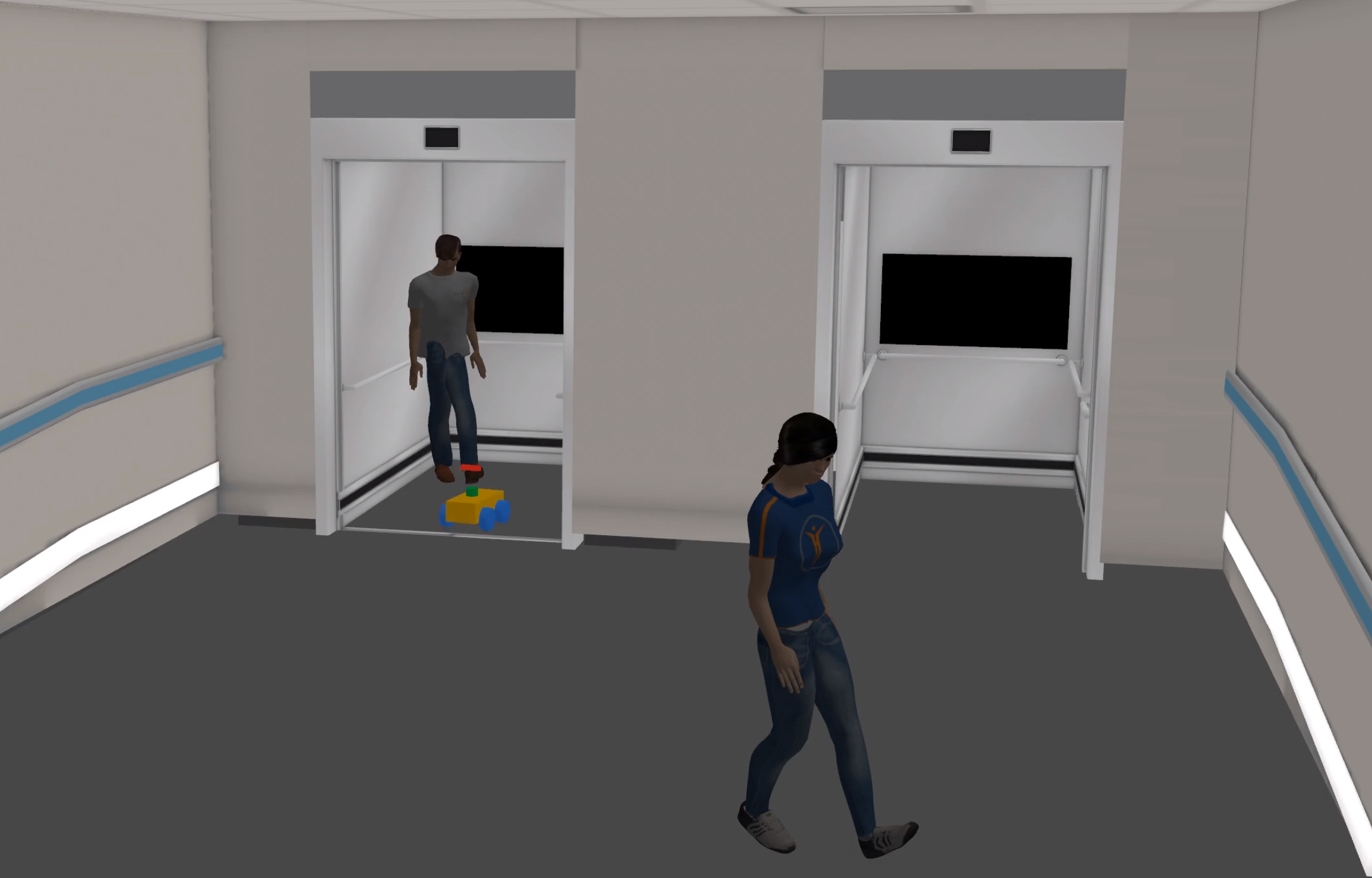}
        \caption{Elevator Scenario}
        \label{fig:res_elevator}
    \end{subfigure}
    \begin{subfigure}{0.47\textwidth}
        \includegraphics[width=0.32\textwidth]{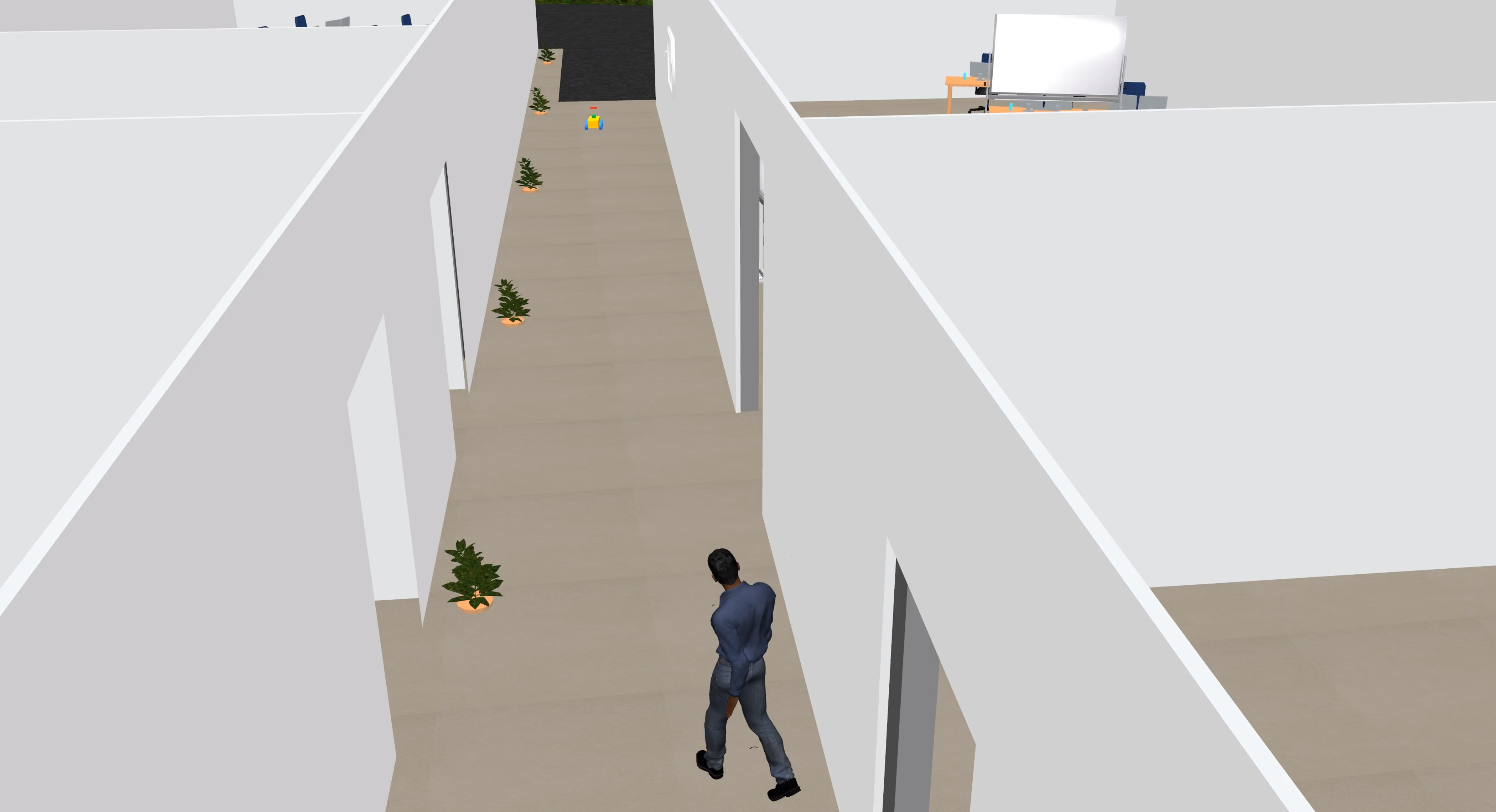}
        \includegraphics[width=0.32\textwidth]{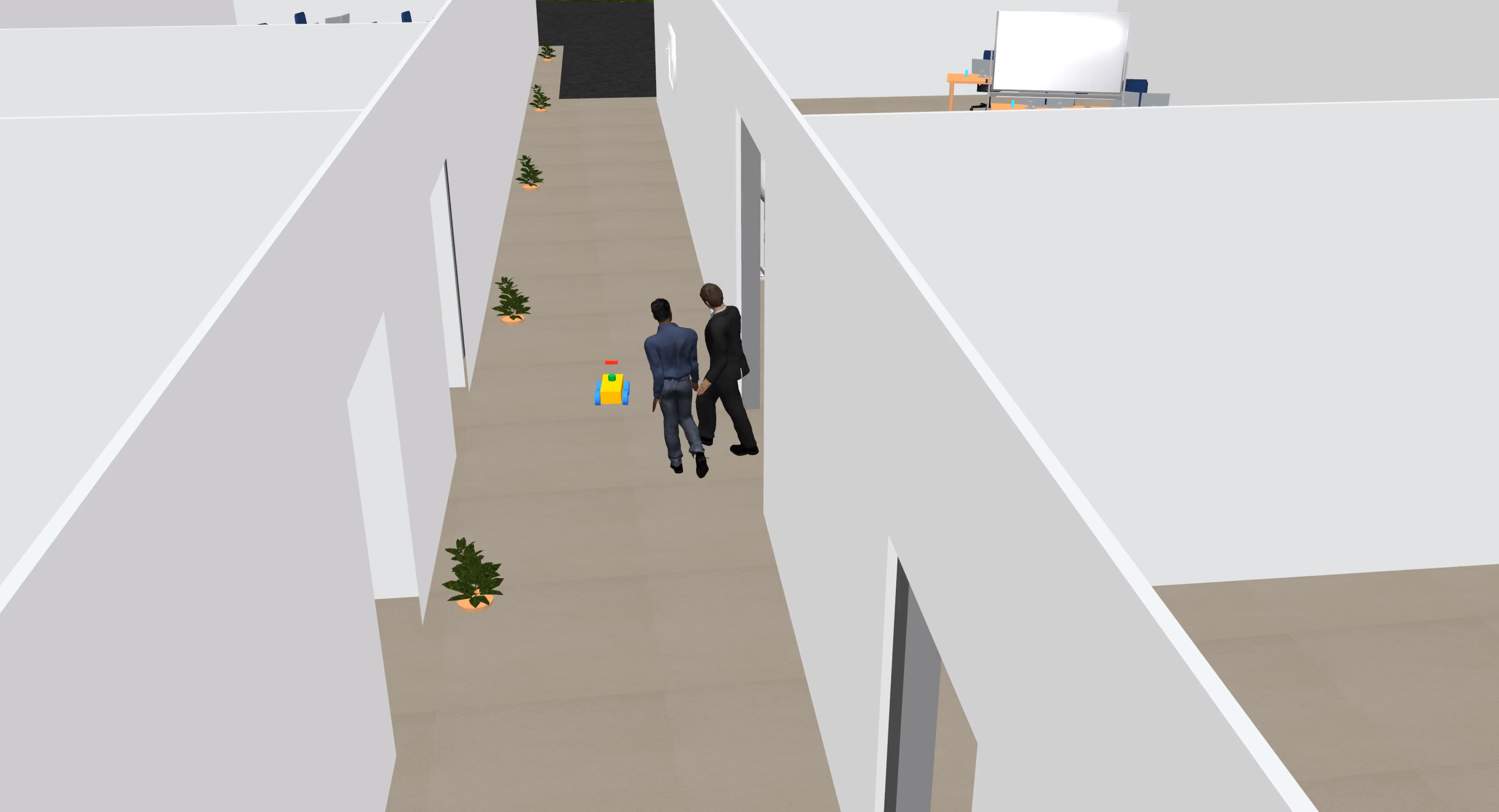}
        \includegraphics[width=0.32\textwidth]{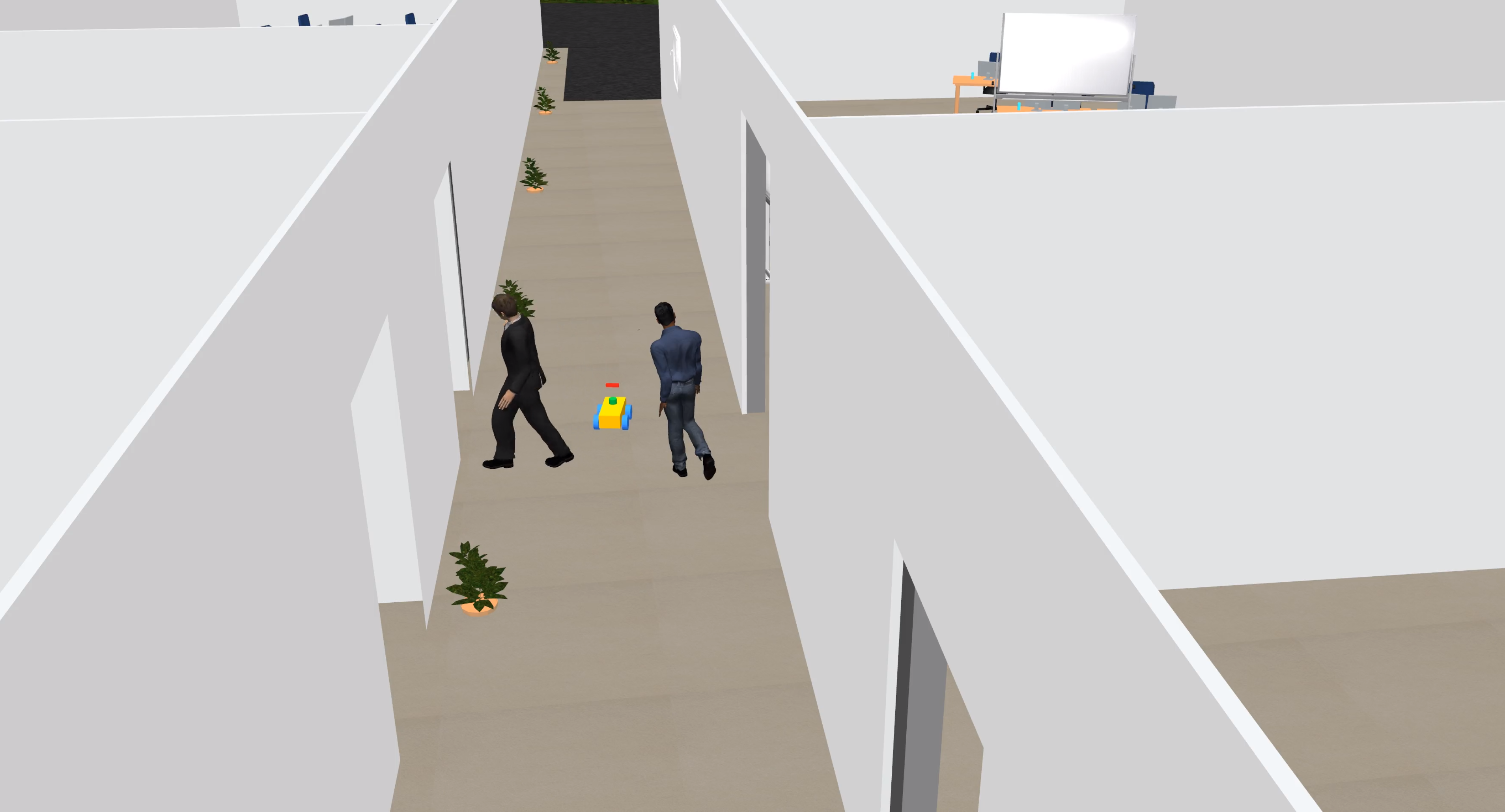}
        \caption{Intersection Scenario}
        \label{fig:res_corridor}
    \end{subfigure}
    \caption{Two of the scenarios generated in the Persona-based Assessment}
    \label{fig:selected_scenarios}

\end{figure}

After generating and visualizing the scenarios in Gazebo, Amy easily recorded videos for use in a questionnaire. Snapshots of the videos for selected scenarios are shown in {Fig. \ref{fig:selected_scenarios}}. She selected questions from the PSI\cite{barchard2020measuring} and added questions about social navigation principles~\cite{francis2023principles}, focusing on predictability, contextual appropriateness, and proactivity.

Amy surveyed 40 participants per scenario (adults with no vision disabilities) through Prolific\footnote{\url{https://www.prolific.com/}} --- note that we actually conducted this study as part of our persona-based assessment. The results revealed significant differences between the algorithms in the scenarios. For example, as shown in Fig.~\ref{fig:qualtrics_scores}, the basic MPPI algorithm received higher Perceived Social Intelligence (PSI) scores than Nav2Can in the corridor scenario, while Nav2Can performed better in the warehouse scenario. This was reinforced by participants comments: ``\textit{I think it} [MPPI] \textit{was very effective as it got through the corridor at a good speed without being too much of an issue for people passing through.}'', ``[Nav2Can] \emph{moved out of the way to go around the group of people.} [MPPI] \textit{and} [HA-MPPI] \textit{drove right into the group of people, which could cause accidents in the warehouse}." This led Amy to suggest that different algorithms might be more suitable for different social contexts, or further development was needed on general social navigation.

To summarize, this study affirmed the findings from our design validation and usability studies. Without SocRATES, we estimate that Amy would have spent approximately a month conducting this study manually, compared to 4 days with SocRATES. The study also revealed areas for improvement, such as in trajectory generation and modification.

 \section{Conclusion}
 In this work, we introduced SocRATES, an automated system for generating diverse, contextually rich social navigation scenarios in simulated environments. Through usability studies with researchers and a persona-based assessment, we demonstrated the system's effectiveness and ease of use. SocRATES marks an important step towards scalable, scenario-based testing and benchmarking for social navigation algorithms. By incorporating human evaluations alongside scenario-based testing, we aim to complement existing proxemics-focused benchmarks and enable the evaluation of subjective, hard-to-define social metrics.

Building on the feedback from our user study, we plan several improvements to enhance the system’s robustness and user experience. First, LLMs do make reasoning errors and we plan to explore fine-tuning models specifically for scenario generation to improve reliability. Additionally, the current requirement for users to manually annotate maps can be time-consuming, so we plan to automate this process and include procedurally generated environments based solely on social context and location descriptions. Finally, we aim to develop a validation module to ensure that the generated scenarios accurately reflect the user’s input. With these enhancements, SocRATES has the potential to significantly advance the testing and evaluation of social navigation algorithms, providing a scalable, flexible platform for assessing both technical performance and social behavior.

 \balance
\bibliography{refs}

\end{document}